\documentclass{article}

\usepackage[preprint,nonatbib]{Styles/neurips_2024}

\usepackage{marvosym}
\usepackage[utf8]{inputenc} 
\usepackage[T1]{fontenc}    
\usepackage{hyperref}       
\usepackage{url}            
\usepackage{booktabs}       
\usepackage{amsfonts}       
\usepackage{nicefrac}       
\usepackage{microtype}      
\usepackage{xcolor}         

\usepackage{bbding}
\usepackage{amsmath}
\usepackage{tcolorbox}
\usepackage{lipsum} 

\usepackage[utf8]{inputenc} 
\usepackage[T1]{fontenc}    
\usepackage{url}            
\usepackage{booktabs}       
\usepackage{amsfonts}       
\usepackage{nicefrac}       
\usepackage{microtype}      
\usepackage{graphicx}
\usepackage{multirow}
\usepackage{pifont}
\usepackage{svg}
\usepackage{graphicx}
\usepackage{booktabs} 
\usepackage{tabularx} 
\usepackage{xcolor}
\newcommand{\modelname}{GUI Narrator~}
\newcommand{\datasetname}{Act2Cap~}
\usepackage{appendix}
\definecolor{citecolor}{HTML}{0071bc}
\DeclareUnicodeCharacter{FF0C}{,}

\definecolor{darkgreen}{RGB}{60,179,113} 
\definecolor{darkred}{RGB}{255,160,122} 

\title{GUI Action Narrator:\\Where and When Did That Action Take Place?}

\author{%
        Qinchen Wu$^1$,~
	Difei Gao$^1$,~
	Kevin Qinghong Lin$^1$,~
	Zhuoyu Wu$^2$,~
        \\
        \textbf{
        Xiangwu Guo$^1$,~ 
        Peiran Li$^1$,~ 
        Weichen Zhang$^1$,~
        Hengxu Wang$^1$,~
        Mike Zheng Shou$^1$\textsuperscript{\Letter}
        }
	\\
	$^1$Show Lab, National University of Singapore\quad
        $^2$Chinese Academy of Sciences, Shenzhen \\
        \url{https://showlab.github.io/GUI-Narrator}
}

\begin{document}

\maketitle

\begin{abstract}
The advent of Multimodal LLMs has significantly enhanced image OCR recognition capabilities, making GUI automation a viable reality for increasing efficiency in digital tasks. One fundamental aspect of developing a GUI automation system is understanding primitive GUI actions. This comprehension is crucial as it enables agents to learn from user demonstrations, an essential element of automation. To rigorously evaluate such capabilities, we developed a video captioning benchmark for GUI actions, comprising 4,189 diverse video captioning samples. This task presents unique challenges compared to natural scene video captioning: 1) GUI screenshots typically contain denser information than natural scenes, and 2) events within GUIs are subtler and occur more rapidly, requiring precise attention to the appropriate time span and spatial region for accurate understanding. To address these challenges, we introduce our GUI action dataset \textbf{\datasetname} as well as a simple yet effective framework, \textbf{\modelname}, for GUI video captioning that utilizes the cursor as a visual prompt to enhance the interpretation of high-resolution screenshots. Specifically, a cursor detector is trained on our dataset, and a multimodal LLM model with mechanisms for selecting keyframes and key regions generates the captions. Experimental results indicate that even for today's most advanced multimodal models, such as GPT-4o, the task remains highly challenging. Additionally, our evaluations show that our strategy effectively enhances model performance, whether integrated into the fine-tuning of open-source models or employed as a prompting strategy in closed-source models. 

\end{abstract}

\section{Introduction}
GUI Automation holds significant importance as it streamlines and optimizes user interactions with Graphical User Interfaces (GUI), drastically improving the efficiency of digital tasks, including information seeking, online shopping, and software copilot. Existing research for this topic can be broadly categorized into two branches: Some works~\cite{chawla2024guide_intro-1,koh2024visualwebarena_intro-1,ma2024comprehensive_intro-1,nass2023improving_intro-1,hong2023cogagent}
 propose new pre-training models to better understand and ground GUI elements~\cite{niu2024screenagent_intro-1,sun2022meta_intro-1,wang2024understanding_intro-1,zhang2024android_intro-1}, while others~\cite{liu2023chatting_intro-2,liu2023fill_intro-2,pan2023autotask_intro-2,wen2024autodroid_intro-2} focus on building AI agents for action generation. However, one crucial capability remains unaddressed for building a powerful GUI automation system: comprehending the specific GUI actions depicted in screenshot recordings. Recognizing these actions not only enables more effective reproduction of user behavior but also provides insights into user engagement with applications. Such capabilities significantly extend the functionality of GUI automation systems, making them more versatile and intelligent in handling real-world tasks.

\begin{figure}[htbp]
    \centering
    \includegraphics[width=1\linewidth]{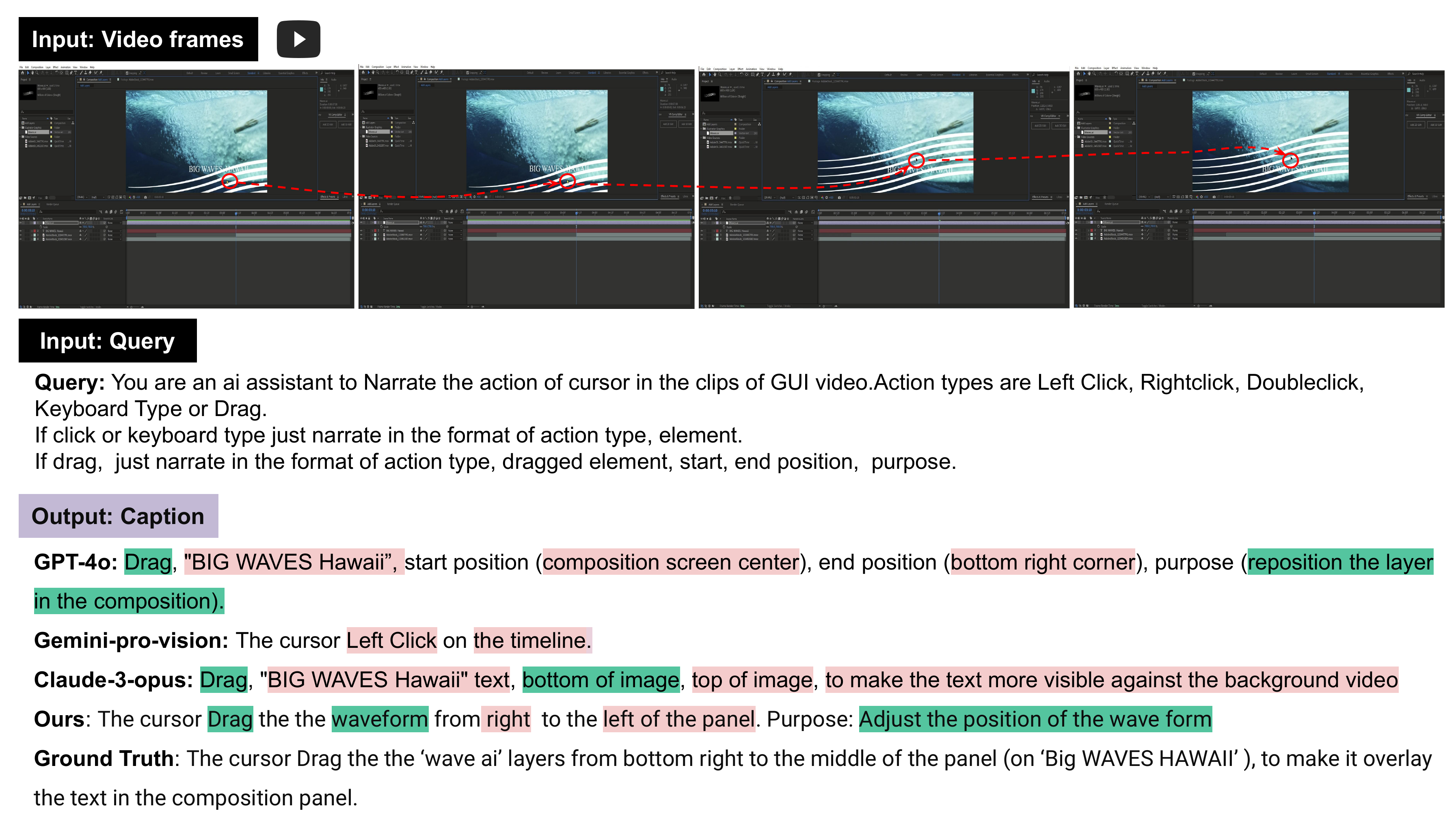}
    \vspace{-0.6cm}
    \caption{\textbf{Illustration of GUI Action Narration}. Comparing the action narration generated with closed-source models and our result. The \textcolor{darkgreen}{green} color indicates correct, while the \textcolor{darkred}{red} indicates wrong.}
    \label{fig:1}
    \vspace{-1cm}
\end{figure}

Understanding the actions in natural scenes (i.e., video captioning)~\cite{seo2022end_intro-3,pan2020spatio_intro-3,lin2022swinbert_intro-3,yang2023vid2seq_intro-3} is a well-established task in multimedia and machine learning. However, the GUI domain introduces a distinct set of challenges that deviate significantly from those encountered in natural settings. First, GUI screenshots contain much denser information than natural scenes, with numerous interactive elements packed into small areas. Second, the actions within GUIs are subtler and occur more rapidly, demanding a more refined temporal resolution to accurately capture fleeting interactions. Third, precise spatial localization is crucial as the exact positioning of the cursor and other elements can dictate the meaning and outcome of an action. Given these specialized demands, it becomes essential to develop a tailored benchmark for GUI video captioning.

To address the unique demands of GUI video captioning, we have meticulously crafted a GUI video caption benchmark, named \datasetname, encompassing 4,189 samples, as shown in Fig.~\ref{fig:1}. Each video in the dataset meticulously records a distinct GUI action, such as left-clicks, right-clicks, double-clicks, drags, and typing, spanning a variety of software environments. This diversity in software types includes Adobe Premiere Pro, Adobe After Effects, Office, and Web Tools, reflecting the broad applicability of GUI automation across different platforms. The AI models tasked with analyzing these videos must not only accurately identify the type of action performed but also the specific interface elements involved, such as buttons, menus, or text fields. The dataset is carefully segmented into over 3,152 automatically captured GUI action videos, which were generated through scripted interactions designed to mimic real-user behavior, and 1，037 manually annotated videos that provide detailed, human-verified labels of GUI actions.

To address the challenges of GUI video captioning, we have developed a simple yet effective framework, named \modelname, tailored specifically for this domain. Our approach centers on using the cursor as a visual prompt, a novel strategy that harnesses the inherent focus point of user interactions in GUIs. To enhance the model’s attention to high-resolution details around the cursor, where actions are most likely to occur, we train a lightweight detection model capable of locating the cursor in various formats. The model then uses the grounded cursor to identify the keyframes where events occur in the video. The framework employs two images to represent a screenshot at a single timestamp: it uses a lower resolution for the broader screen area and processes the area around the cursor in high-resolution to maintain context while conserving computational resources. Ultimately, a multimodal LLM is implemented to generate captions based on this new representation of keyframes.

We conducted extensive evaluations of leading open-source and closed-source models on our newly developed benchmark to understand their capabilities and limitations in GUI video captioning. These evaluations revealed that even the most advanced models struggle with the unique demands of GUI scenarios. Notably, the best-performing model, GPT-4o~\cite{openai2024gpt4o}, achieved only a 19.5\% accuracy, highlighting the challenges posed by dense information and rapid action sequences typical of GUI interactions. Furthermore, our results also showcased the effectiveness of our framework. By integrating our strategies, both open-source and closed-source models showed marked improvements in performance. This suggests that our approach not only addresses the immediate challenges but also provides a scalable solution that enhances the adaptability and accuracy of models in handling complex GUI environments.

\section{Related Work}

\textbf{GUI-Related Benchmark.}
In the field of GUI agents for task automation, current benchmarks can be broadly categorized into two distinct types: UI-grounded benchmarks and UI automation benchmarks. UI-grounded benchmarks~\cite{Mind2Web, UGIF_arxiv, seeclick} assess the ability of agents to understand and interpret the graphical user interface based on visual cues and natural language commands, focusing on the agent's proficiency in mapping these inputs to specific UI elements and actions. On the other hand, UI automation benchmarks~\cite{Mind2Web, WebShop_Yao2023, gao2023assistgui, YOLA_MM, kim2024languageNIPS,zhang2023reinforced} are designed to evaluate the performance of these agents in executing complex sequences of actions within an interface, thereby measuring their capability to automate tasks effectively across different applications and platforms. 
Differing from previous works~\cite{seeclick,gpt4vinwonderland,hong2023cogagent,yang2023appagent}, ours introduces the GUI Action Narrator benchmark, which aims for models to understand GUI recordings. This approach serves as a complement to existing GUI agents, enabling models to grasp the meaning of user demonstrations. It enhances the model's comprehension of the functions of UI elements and, in the future, could assist models in learning automation knowledge directly from user demonstrations.

\textbf{Video Captioning Dataset.} The domain of video captioning is supported by a diverse array of datasets that cater to different scenarios. The MSVD~\cite{MSVD} and YouTube2Text~\cite{Youtube2textguadarrama2013y} focus on general activities, while MSR-VTT~\cite{MSR-VTT} expands the scope with a broader range of video-text pairs. Activity recognition is addressed by datasets such as ActivityNet~\cite{Activitynet_Caba} and Charades~\cite{Charades_sigurdsson2016hollywood}, with Charades-Ego~\cite{Charades-ego} providing a first-person perspective. For fine-grained action recognition, the Something-Something V2 dataset~\cite{Something-Something} stands out. Some other works focus on plot understanding. The MPII-MD~\cite{MPII-MD_rohrbach2015dataset} and LSMDC~\cite{LSMDC_rohrbach2017movie} datasets offer a cinematic perspective with video clips from movies. More recently, HowTo100m provides a large collection of instructional videos, and cooking-specific datasets like YouCook II~\cite{YouCook2_zhou2018towards} and TACoS-Mlevel~\cite{TACOS-Mlevel_rohrbach2014coherent} cater to the culinary domain.
Previous video caption datasets have primarily focused on natural scenes, whereas our work is dedicated to understanding videos of GUI actions. This represents a completely different domain from natural scenes and presents unique challenges, including capturing detailed GUI information and quickly occurring actions.

\textbf{Multi-modal Large Language Model.} In the domain of Multimodal Large Language Models (MLLMs), several pioneering models~\cite{MLLMs_survey} have made significant strides. BLIP-2~\cite{BLIP2_li2023}, Llava~\cite{Llava_liu2024}, and Video-LLaMa~\cite{Video-LLama}, etc. integrate a strong visual encoder with a language model, demonstrating effectiveness on a range of visual-language tasks. These models, along with others like Qwen-VL\cite{Qwen_technicalReport,bai2023qwen}, which adeptly handles interleaved image-text data, bring a new dimension to MLLMs' capabilities. Later developments have broadened these applications and functionalities~\cite{gao2023assistgpt,yang2023mm}. This includes enhanced granularity where refined control over the user's visual prompts allows for targeting specific regions using bounding boxes or selecting particular objects with a click. There has also been a significant expansion in supporting OCR-related understanding, such as document parsing~\cite{Donutparser, rausch2021docparser, shen2021layoutparser}, GUI understanding~\cite{fuyu-8b,hong2023cogagent,seeclick}. 
Our approach is unique as it represents the early attempt to understand GUI Action Videos. It leverages the cursor as an inherent visual prompt to assist with high-resolution, text-heavy, and fast-action screenshot recordings.

\section{\datasetname Benchmark}

\subsection{Task Formulation}
In this section, we introduce \datasetname. The video narration for atomic actions can be formulated as shown in Fig.~\ref{fig:1}. The input for the VLM model will be a full-resolution screenshot recording, accompanied by a prompt to guide the MLLM in generating the corresponding caption. The output of the model is a natural language caption.


\subsection{Data Collection}
\begin{figure}[t]
    \centering
    \includegraphics[width=1\linewidth]{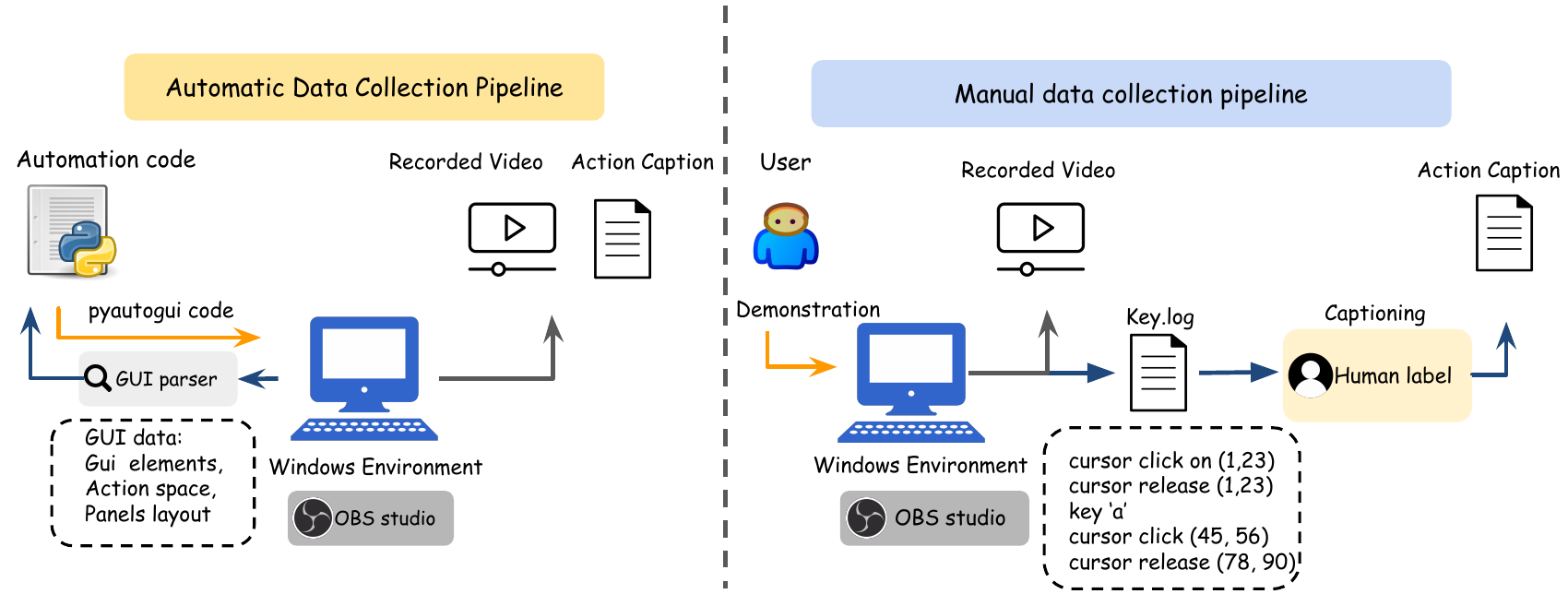}
    \vspace{-0.6cm}
    \caption{\textbf{Data Collection Pipeline}: (left) Our automatic data collection pipeline and (right) manual data collection pipeline.}
    \label{fig:2}
\end{figure}

Our dataset consists of a wide range of GUI actions covering cursor actions (including Left-Click, Right-Click, Double-Click and Drag) and keyboard Type actions. 
The data collected for GUI Narrator consists of two separate pipelines shown in Fig.~\ref{fig:2}. One is the action videos and annotations collected by automatic environment and the other is collected by the human demonstration method. Both datasets are collected under the Windows GUI environment.

\textbf{Video Collection.}
The videos in our benchmark are collected through two distinct pipelines using OBS Studio to capture the complete action video. Initially, our training datasets are generated via an \textbf{automatic data collection pipeline}. A GUI parser \cite{gao2023assistgui} first analyzes the GUI layout into three main components: 1) Interactive GUI elements; 2) Action spaces such as \texttt{Left-Click, Right-Click, Double-Click, Drag,} and \texttt{keyboard typing} for each element; 3) Coordination of specific panels like the \texttt{timeline, component,} and \texttt{effect panels}, which are crucial for determining the starting or ending positions of the \texttt{Drag} action. For actions such as Click or Type, the automatic data collection pipeline executes \textit{pyautogui} code based on the selected [element, action type] pair and the parsed button locations. For Drag actions, the pipeline generates execution code tailored to [element, action type, panels] to direct the cursor in performing drag operations. To emulate human interaction in GUI videos, the starting time of the cursor is randomized, thereby avoiding the introduction of rigid temporal knowledge regarding the start and end times of the atomic action video.

Although the automatic data collection pipeline efficiently scales up data collection, it struggles with executing more complex, human-like actions such as \texttt{dragging the cursor to draw a circle}. To address these challenging tasks, we introduced a second pipeline known as the \textbf{manual data collection pipeline}. Similar to the automatic data collection pipeline, user-demonstrated videos are recorded using OBS Studio and are subsequently segmented into sub-videos based on the recorded key logs, with each sub-video containing a single action.

\begin{table}[b]
\centering
\resizebox{\textwidth}{!}{%
\label{tab:dataset_comparison}
    \begin{tabular}{lccccccc}
    \toprule
    \multirow{2}{*}{\textbf{Datasets}} & \multirow{2}{*}{\textbf{Data size}}  &\multirow{2}{*}{\textbf{Task domain}}& \multicolumn{2}{c}{\textbf{Visual components}} & \multicolumn{2}{c}{\textbf{Label components}} \\ 
    \cmidrule(lr){4-5} \cmidrule(lr){6-7}
    & & &\textbf{Cursor} & \textbf{Format} &\textbf{Coordinate} & \textbf{Action Caption}  \\ 
    \midrule
    ScreenSpot~\cite{seeclick} & 600 & Grounding &{\XSolidBrush}  & Single Screenshot & \CheckmarkBold  &{\XSolidBrush}  \\ 
    AgentStudio~\cite{zheng2024agentstudio} & 227 &No supported benchmark task & \CheckmarkBold & Video &\CheckmarkBold & \CheckmarkBold  \\ 
    OSworld~\cite{xie2024osworld} & 369 &Planning  &{\XSolidBrush}  & Screenshots &\CheckmarkBold & {\XSolidBrush}   \\ 
    MIND2WEB~\cite{Mind2Web} & 2,000 &Planning  &{\XSolidBrush}  & Screenshots &\CheckmarkBold & {\XSolidBrush}   \\
    \midrule
    \textbf{Act2Cap (Ours)} & \textbf{4,189}&Captioning & \CheckmarkBold & Video& \CheckmarkBold & \CheckmarkBold \\ 
    \bottomrule
    \end{tabular}
    }
\caption{\textbf{Comparison of GUI datasets}: Comparison of the existing GUI dataset on the dataset size, task domain, visual components, and label components.}
 \label{tab: dataset}

\end{table}

\textbf{Caption Annotation.}
The caption annotation process varies between the automatic data collection pipeline and the manual data collection pipeline.
In the automatic collection pipeline, captions are generated based on the predefined action space. For example, given a specific element and action, the annotation would be formatted as [\texttt{Specific Action} on \texttt{element}] or [Drag the \texttt{element} from \texttt{start position} to \texttt{end position} for \texttt{Purpose}].
For the manual collection pipeline, user actions are logged in the log file, which records the time each action occurred and the position of the mouse click. Screenshots are taken based on these timestamps to document the state before and after the action. Then, we invite annotators to label these actions based on the captured screenshots.

\textbf{Datasets Setting.}
We constructed our training set using 3,152 video narration pairs collected from the automatic data collection pipeline, complemented by 488 high-quality data samples from human demonstrations to address actions that are challenging for the automatic pipeline to capture (e.g., \textit{drawing shapes, adjusting the size or orientation of elements}). Additionally, we carefully selected 549 video narration pairs from human demonstrations, ensuring no overlap between the human demonstration data split and the test set.
Furthermore, the test dataset includes GUI panels from environments such as Web (Amazon, CivitAI, Spotify) and Microsoft Office (Word, Excel), which are not present in the training dataset. This inclusion makes our GUI action captioning benchmark more challenging.

\textbf{Statistics.}
The following Fig.~\ref{fig:demo} demonstrates the distribution training dataset and the testing dataset due to the difficulty of scaling up the Drag and Keyboard Type actions. Most of the training datasets are collected under the GUI layout of Adobe Premier Pro and Adobe After Effects. However, the testing dataset comprises additional GUI environments including Word, Excel, and Web environments. Compared to other existing datasets shown in Tab.~\ref{tab: dataset}, our dataset prevails in having more data pairs and action captioning pairs.

\begin{figure}[t]
     \centering
        \vspace{-0.1cm}
	\includegraphics[width=1.0\linewidth]{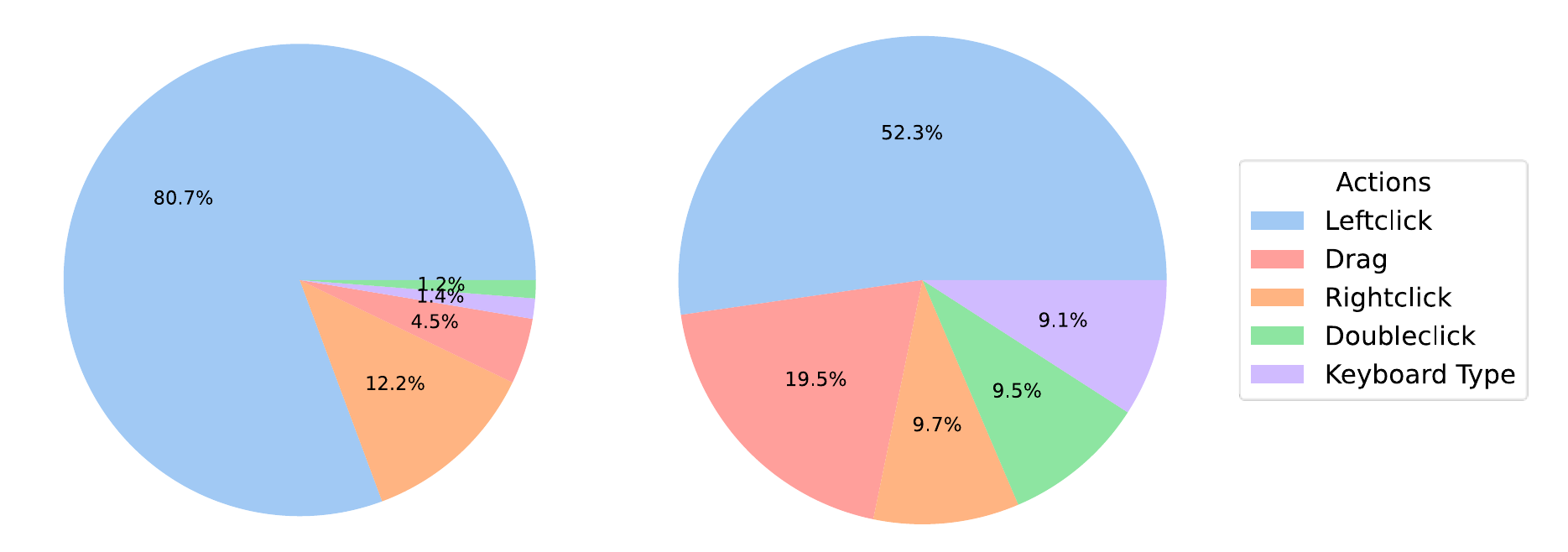}
	\caption{\textbf{Action distribution in training and test dataset}. The left-hand side shows the distribution of training data and the right-hand side demonstrates the test dataset.}
\label{fig:demo}
\end{figure}

\subsection{Metrics}
Assessing the performance of narration should be based on the semantic meaning of the elements in sentences. Evaluating correctness on a word-by-word basis lacks the ability to capture the understanding of the overall semantic content. To address this issue, we employ the GPT-4 language model as LLM evaluator to extract key elements from the narrations and then assess whether each element matches the semantic meaning in element-wise, assigning a score of 0 for different and 1 for same. Because the elements contained in Click, Keyboard Type and Drag narrations are different, we categorize actions into three types for evaluation: Click actions (including Left-Click, Right-Click and Double-Click), Drag actions, and Keyboard Type actions, as shown in Tab.~\ref{tab:metic}. 

 \textbullet \hspace{1em} \textbf{Click actions}: These are instantaneous actions evaluated based on the name of \texttt{GUI element} and the type of cursor click.
 
\textbullet \hspace{1em} \textbf{Drag actions}: These require consideration of 4 factors: \texttt{the start position}, \texttt{end position}, \texttt{GUI element} involved, and the intended \texttt{purpose} of the drag.

\textbullet \hspace{1em} \textbf{Keyboard Type actions}: These are evaluated based on the \texttt{elements} (i.e. keys) typed or pressed.

\begin{table}[h]
    \small
  \centering
  \begin{tabular}{c c c c c c }
    \toprule
      \textbf{Action type}& \textbf{Specific action} & \textbf{Start} & \textbf{End} & \textbf{GUI element} & \textbf{Purpose}  \\
    \midrule
    Click     & 0 / 1  & - & - & 0 / 1 & - \\
    Keyboard Type  & 0 / 1  & - & - & 0 / 1  &- \\
    Drag      &0 / 1  &0 / 1 &0 / 1 &0 / 1 & 0 / 1\\
    \bottomrule
  \end{tabular}
\caption{\textbf{Elements} to be assessed in predicted captions for each action type in \datasetname.}
    \label{tab:metic}
\end{table}

The LLM evaluator will generate a list of 5 elements for Drag actions and 2 elements for Click and Type actions. To enhance the robustness of this evaluation, we predefined knowledge for the LLM evaluator, considering 'buttons' and 'icons' as well as 'folders' and 'files' to have equivalent semantic meanings. The evaluation score for \modelname is determined by the Intersection over Union (\textit{IoU}), calculated as \(\text{IoU} = \frac{\sum_{i} |\text{P}_i \cap \text{G}_i|}{\sum_{i} |\text{P}_i \cup \text{G}_i|} \), where $i$ denotes the $i_{th}$ value of the justified element index. $P$ represents the model's predictions, while $G$ denotes the ground truth. Specifically, $P \cap G$ indicates the number of elements correctly predicted by the model that matches the ground truth, and $P \cup G$ represents the total number of elements present in either the prediction or the ground truth. 




\section{Methods}
\label{headings}

\begin{figure}[h]
	\centering

\includegraphics[width=\linewidth]{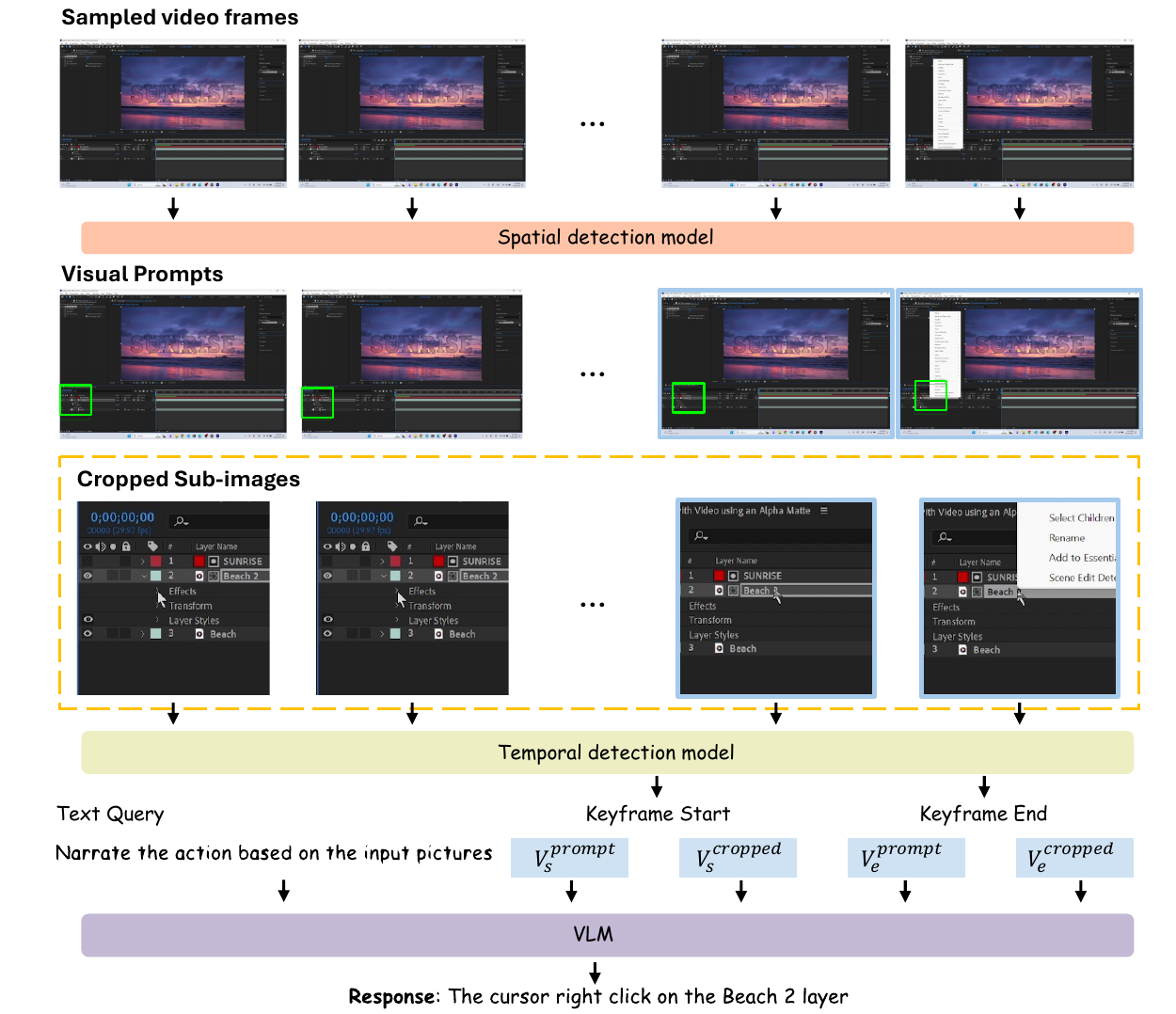}
\caption{\textbf{Overview of GUI Narrator}: It first processes sampled frames from the video through a spatial detection model, which locates the cursor, adds visual prompts to the screenshot, and crops the region near the cursor to represent each frame. Subsequently, the temporal detection model identifies further keyframes based on the cropped sub-images. The extracted keyframes, combined with a text query, are then fed into the VLM model, which generates a narration describing the GUI actions.}
\label{fig:overview_baseline}
\end{figure}

\textbf{Overview.} We propose a two-stage baseline approach for atomic step narration in graphical user interfaces (GUIs) as depicted in Fig.~\ref{fig:overview_baseline}. In the initial stage, the input video is sampled uniformly into 10 frames. A trained cursor location detector will be utilized to detect the cursor in each sampled frame. The keyframes representing the GUI screenshot before and after atomic action will be extracted by the temporal detection model. The second stage takes the two keyframes extracted from the first stage as well as visual prompt annotation and cropped regional images as the visual input query for a closed-source VLM model or our fine-tuned open-source model QwenVL-7B~\cite{Qwen_technicalReport}. The output of the model will be the action narration containing the \texttt{Action type}, \texttt{Element}, \texttt{Purpose}.

\subsection{Action-aware Spatial Sampling}
\textbf{Visual Prompt Format.}
The use of visual prompts, such as colored bounding box annotations on screenshots, offers a promising method to improve the understanding of GUI panels. The current baseline approach, which uses the entire image as input, makes it difficult for vision models to focus on the action region. In contrast, our visual prompts employ green bounding boxes of fixed size directly annotated on the original screenshot images, thereby forcing the model to pay attention to the regional information around the cursor limited by the bounding box. Moreover, we cropped out the region of the screenshot as a high-resolution input to demonstrate the detailed information for the following stages.

\textbf{Visual Prompt Creation.}
The coordination of the bounding box is generated based on the detected center location of cursor $c=(x, y)$, width and height of the image ($w_i, h_i$), and the size of the bounding box ($s_{box}$). We trained our cursor detection model on YOLO-v8 \cite{redmon2016you} model under our customized dataset comprising image-cursor pairs. Given the video frame list of [$V_1$, $V_2$,... ,$V_{N}$], where $V_i$ denotes the $i_{th}$ of sampled frame, the output will be [$V_{1}^{prompt}$, $V_{2}^{prompt}$,..., $V_{N}^{prompt}$] and [$V_{1}^{cropped}$, $V_{2}^{cropped}$,..., $V_{N}^{cropped}$], where $N$ is the number sampled frame. The $prompt$ and $cropped$ denote the screenshot with printed visual prompts and the cropped sub-images from the sampled frame $V$.

\begin{figure}[h]
	\centering
	\includegraphics[width=1 \linewidth]{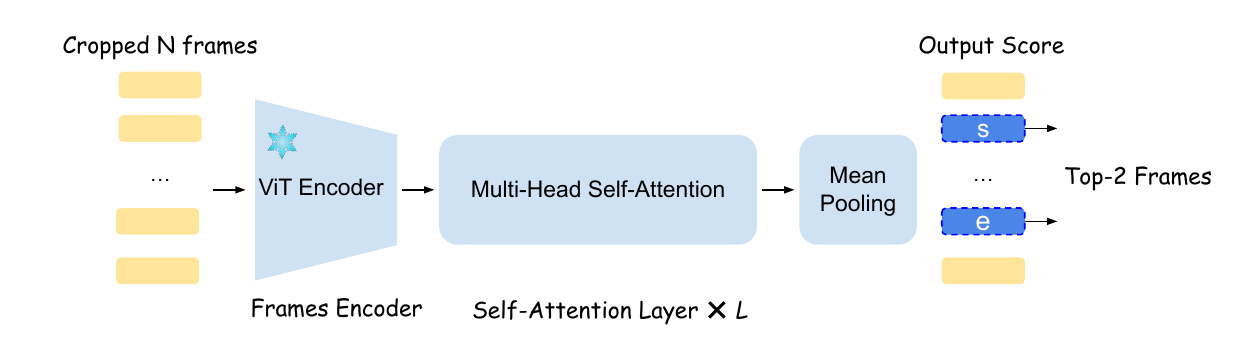}
	\caption{\textbf{Temporal Detection Model}: We implement a frozen ViT encoder from OpenCLIP pre-trained model together with trainable Multihead-Self-Attention Layers. }
 
\label{fig:temporal}

\end{figure}

\subsection{Action-aware Temporal Sampling}
\textbf{Frames Encoder.}
The temporal prompt of the video shown in Fig.~\ref{fig:temporal} aims to extract the 2 frames that represent the GUI screenshot before and after the action among the $N$ cropped images around the cursor denoted as [$V_{1}^{cropped}$, $V_{2}^{cropped}$,... $V_{N}^{cropped}$]. We utilize the pre-trained visual encoder from CLIP to encode the images. Suppose the encoding dimension is $d_v$, the output feature of the Frames Encoder will be of shape $N \times d_v$.

\textbf{KeyFrame Extractor.}
To identify the keyframes that most accurately depict the GUI screenshot before and after the action, we employ self-attention layers to capture the interdependencies among the frame representations. The resulting model output is a tensor of dimensions $N \times 1$, from which the top-2 highest value indicates the keyframes denoted as $s$ and $e$.

\subsection{Action Captioning}
We fine-tuned an open-source model QwenVL-7B~\cite{Qwen_technicalReport} for action captioning. The visual inputs consists of a list of image [$V_{s}^{prompt}$, $V_{e}^{prompt}$, $V_{s}^{cropped}$, $V_{e}^{cropped}$]. The text input is designed to guide the model to not only recognize the entire screenshot but also to focus on the region of interest that is annotated and cropped separately. A template of the text query is: \emph{The cursor is located in the annotated green bounding box. The third and fourth image shows the cropped detailed image around the cursor before and after the action.}  The learning objective of the VLM model is to minimize the loss of the model $\mathcal{L}$ on model parameters $\Theta$. $u_i$ denotes the word being predicted and $k$ represents the context window.
\begin{equation}
   \mathcal{L} = \sum_{i} \log p(u_i \mid u_{i-k}, \ldots, u_{i-1}; \Theta).
\end{equation}

\section{Experiments}
\subsection{Implementation Details}
We train our models on A5000 NVIDIA GPUs in two stages. In the first stage, we train the cursor detection model using cursor annotations in our dataset to establish cursor grounding. Following this, we train the KeyFrame Detection model based on the cursor grounding results. In the second stage, we freeze the model from the first stage and focus solely on training the QwenVL-7B model with our proposed baseline \modelname (7B). A similar mechanism can also apply to a closed-source MLLM with zero-shot inference. We build one version upon GPT-4o, denoted as \modelname(GPT-4o).


\subsection{Comparison with State-of-the-arts}
The following Tab.~\ref{tab:SOTA} shows the score of the most popular VLM models that support multiple image input. The models are provided with original screenshots and the start frame of the video and the end frame of the video.

\begin{table}[tbp]
  \centering

  \resizebox{\textwidth}{!}{%
 \vspace{-0.1cm}

  \begin{tabular}{lc c c c c| c}
    \toprule
        \textbf{Base model} & \textbf{Left-Click} & \textbf{Double-Click} & \textbf{Right-Click} & \textbf{Drag} & \textbf{Keyboard Type} &  \textbf{Average score}\\
    \midrule
    \ Gemini-pro-vision\cite{team2023gemini} & 48.8 &2.4 &3.7 &0.56 &2.7 &11.6    \\
    \ Claude-3-opus\cite{anthropic2024claude} &26.8 &3.4  &8.4  &21.1 &\textbf{35.1} &18.9  \\
    \ GPT-4o\cite{openai2024gpt4o} &30.0 & 4.6 &18.1 &17.3 &27.5 &19.5   \\
    \ \textbf{\modelname (GPT-4o)}\cite{openai2024gpt4o} &\textbf{52.6} & \textbf{21.1} &\textbf{29.4} &\textbf{21.4} &34.7 &\textbf{31.8}    \\
    \midrule
    \ QwenVL-7B (finetuned)\cite{bai2023qwen} &45.2  &0.3 &1.7 &0.17 &1.3 &9.73    \\
    \ \textbf{\modelname (7B)} &\textbf{60.3} &\textbf{7.9} &\textbf{6.4} &\textbf{9.1} &\textbf{33.8} &\textbf{23.5}    \\
    \bottomrule
  \end{tabular}
  }
  \vspace{-0.1cm}
  \caption{Caption Score (\%) of state-of-the-art closed-source and open-source MLLMs.}
    \label{tab:SOTA}
\vspace{-0.6cm}
\end{table}

GPT-4o outperforms others in the average score, with its highest score of recognizing the \texttt{Double-Click} and \texttt{Right-Click} by 4.6 and 18.1 respectively. Although the model reached the highest score of Double-Click among others, it struggled to recognize this kind of task. Claude 3 opus model performs well on \texttt{Drag} actions and \texttt{Keyboard Type} actions. However, it struggled with the captioning the action of \texttt{Double-Click} and \texttt{Right-Click}. For Gemini-pro-vision, despite achieving the highest score for the \texttt{Lef-Click} action, it struggles to differentiate this action from others. In our experiments, it frequently misclassified various actions as \texttt{Left-Click}, resulting in lower scores for the other four action types.

Our proposed \modelname (7B)~suppresses the GPT-4o in Average score. The result also demonstrated that although the trained model performs better on the \texttt{Left-Click} action and \texttt{Keyboard Type} action, it still has difficulty in dealing with \texttt{Double-Click},\texttt{Right-Click} and \texttt{Drag} action. \modelname (GPT-4o) further improves the performance with a large margin.

\subsection{Ablation Study}

\textbf{Ablation on Visual Prompt Size.}
The selected size for the visual prompt plays a critical role in this methodology, as it defines the area on which the model can concentrate. Tab.~\ref{tab:size} presents the scores of the GUI Action Narrator for visual prompt sizes of {128}, {256}, {512}, and {768}. In this ablation study, frames were chosen based on timestamps from the key.log file recorded during human demonstrations, simulating the optimal output of the temporal attention model. The table reveals that the highest average score is achieved with a size of {256}. While reducing the visual prompt size can help the model generate more precise captions for elements such as buttons and icons, it limits the region of interest in the image, which may negatively impact action categorization.

\begin{table}[htbp]
  \centering
  \vspace{-0.3cm}
  \begin{tabular}{ccc}
    \toprule
    \textbf{Visual Prompt Size} & \textbf{Base model} & \textbf{Score}\\
    \midrule
    \ 128 &  & 23.7    \\
  \ 256 & \modelname(7B) & \textbf{25.8}    \\
    \ 512 &  &21.2   \\
    \ 768 &  & 17.4   \\
    \bottomrule
  \end{tabular}
    \caption{Captioning Score~(\%) of GUI Narrator (7B) with different sizes.}
      \label{tab:size}
    \vspace{-0.4cm}
\end{table}

\textbf{Ablation on Spatial Prompt}
We analyzed the methods of spatial prompting in the first stage of this experiment. The frames were selected based on timestamps from the key.log file to simulate the optimal output of the temporal attention model. The evaluation scores shown in Tab.~\ref{tab:visual} indicate that providing annotated prompts on the original images improves the overall performance of the VLM model by conveying the cursor's location, which the model might otherwise miss.

For the GPT-4o model, we input the entire image as a visual query if only one of the visual prompt components is applied. When taking both of the visual components as visual query, we resize the original screenshot with the visual prompt to dimensions of 960$\times$512. This adjustment results in a reduction in the number of visual tokens from 1105 to 680, making the latter method with two visual prompt components more efficient in terms of token utilization compared to the former.

Interestingly, using visual prompts for GPT-4o negatively affects its ability to narrate Right-Click actions. This can be attributed to the failure to detect the cursor's location, which misguides the GPT-4o model's attention within the GUI panel. Conversely, providing cropped images resulted in a higher score for the \texttt{Drag} action compared to using both annotated and cropped images. This is because, without the general screenshot information, the model is more likely to classify the action as a \texttt{Drag} action.

For the fine-tuned open-source model, providing cropped images along with visual prompts on the original screenshot significantly improved the QwenVL-7B model's overall score. This improvement is due to the input size of the QwenVL image encoder being 448 $\times$ 448, which compresses and loses much detail when resizing the original GUI screenshot resolution from 1920 $\times$ 1080. Cropped images provide the model with detailed information about the GUI panel around the cursor, enhancing the model's captioning ability after training. This approach shows the necessity of our Spatial prompt component in our new baseline.

\begin{table}[tbp]
  \centering
  \resizebox{0.95\textwidth}{!}{%
  \begin{tabular}{cccccccc|c}
    \toprule
    \multicolumn{2}{c}{\textbf{Spatial prompt component}} & \multirow{2}{*}{\textbf{Base model}} & \multirow{2}{*}{\textbf{Left-Click}} & \multirow{2}{*}{\textbf{Double-Click}} & \multirow{2}{*}{\textbf{Right-Click}} & \multirow{2}{*}{\textbf{Drag}} & \multirow{2}{*}{\textbf{Type in}} & \multirow{2}{*}{\textbf{Score}} \\
    \cmidrule{1-2}
    \textbf{Visual prompt} & \textbf{Cropped image} \\
    \midrule
    {\XSolidBrush} & {\XSolidBrush} & \multirow{4}{*}{\modelname (GPT-4o)} & 34.8 & 4.9 & 21.5 & 17.5 & 25.5 & 20.8 \\
    \CheckmarkBold & {\XSolidBrush} & & 38.6 & 6.1 & 18.1 & 21.4 & 27.5 & 22.3 \\
    {\XSolidBrush} & \CheckmarkBold & & 63.2 & 22.5 & 27.6 & \textbf{36.5} & 31.2 & 36.2 \\
    \CheckmarkBold & \CheckmarkBold & &\textbf{63.7} & \textbf{26.6} & \textbf{31.0} & 33.1 & \textbf{38.4} & \textbf{38.6} \\
    \midrule
    {\XSolidBrush} & {\XSolidBrush} & \multirow{4}{*}{\modelname (7B)} & 45.2 & 0.3 & 1.8 & 0.17 & 1.3 & 9.8 \\
    \CheckmarkBold & {\XSolidBrush}& & 50.1 & 1.7 & 2.9 & 1.1 & 33.5 & 17.9 \\
    {\XSolidBrush} & \CheckmarkBold & & 63.5 & 4.2 & 12.0 & 7.1 & 31.4 & 23.6 \\
    \CheckmarkBold & \CheckmarkBold & & \textbf{63.9} & \textbf{8.4} & \textbf{12.1} & \textbf{10.0} & \textbf{34.6} & \textbf{25.8} \\
    \bottomrule
  \end{tabular}
  }
 \vspace{0.02cm}
  \caption{Caption Scores(\%) of the VLM models with different combinations of Spatial prompt}
    \label{tab:visual}
 \vspace{-0.5cm}
\end{table}

    
\textbf{Ablation on Temporal Prompt.}
We also analyzed the use of the keyframe extraction model in the temporal grounding stage. We have designed two variants: one uses the first and last frames of the video as keyframes, denoted as \emph{Video Start \& End}; the other uses ground-truth keyframes, denoted as \emph{GT-Keyframes}.
The visual inputs are screenshots with visual prompts and cropped images based on the detected cursor location. The results are shown in Tab.~\ref{tab:temp}. For the GPT-4o model, implementing keyframes improves performance in \texttt{Left-Click} and \texttt{Double-Click} actions, while reducing the scores for \texttt{Drag} and \texttt{Right-Click} actions. This is because Double-Click actions often cause significant changes in the GUI layout, making the keyframes detection model more sensitive to these changes and thus more accurate in detecting keyframes compared to other actions.
A similar phenomenon is observed with our fine-tuned QwenVL model, which shows subtle improvements in both \texttt{Left-Click} and \texttt{Type in} actions but a lower score in \texttt{Right-Click} actions. However, since the QwenVL model is trained on our dataset, it performs better on \texttt{Drag} actions. 
The average score demonstrates an improvement of \textbf{4.6\%} and \textbf{3.9\%} for the GPT-4o model and QwenVL-7B model, respectively, compared to using the start frame and end frame for temporal detection. This improvement is less significant than the improvement observed with the introduction of spatial prompts.

\begin{table}[htbp]
  \centering
  \resizebox{\textwidth}{!}{%
  \begin{tabular}{ccc c c c c |c }
    \toprule
    \textbf{Temporal attention scheme} & \textbf{Base model} & \textbf{Left-Click} & \textbf{Double-Click} & \textbf{Right-Click} & \textbf{Drag} & \textbf{Type in}  & \textbf{Score}                \\
    \midrule
    \ Video Start \& End   & \multirow{3}{*}{\modelname(GPT-4o)} &48.3 &9.4 &\textbf{31.5} &27.3 &35.4  & 30.4    \\
    \ GT-Keyframes  & &\textbf{63.7} &\textbf{26.6} &31.0 &\textbf{33.1} &\textbf{38.4}  &\textbf{38.6}   \\
    \ Temporal Detection Model  &  &52.6 &21.1 &29.4 &21.4 &34.7  & 31.8    \\
    \midrule
    \ Video Start \& End  & \multirow{3}{*}{\modelname(7B)} &59.4 &3.6 &\textbf{12.6} &5.2  &32.1  & 22.6 \\
    \ GT-Keyframes &  &\textbf{63.9} &\textbf{8.4} &12.1 &\textbf{10.0} &\textbf{34.6}  & \textbf{25.8}   \\
    \ Temporal Detection Model  &  &60.3 &7.9 &6.4 &9.1 &33.8 &23.5  \\
    
    \bottomrule
  \end{tabular}
  }
  \vspace{0.02cm}
  \caption{Caption Scores(\%) of the VLM model with ablation of temporal grounding model}
    \label{tab:temp}
  \vspace{-0.7cm}
\end{table}

\section{Conclusion and Limitation}
In this paper, we propose a GUI Video Caption benchmark, featuring 4,189 samples, that addresses unique challenges specific to GUIs, such as denser information and rapid, subtle events. A new captioning framework is also proposed to utilize a cursor as a visual prompt to enhance the interpretation of screenshots. Despite the task's complexity, challenging even for advanced models like GPT-4o, \modelname effectively improves performance. This is demonstrated both in the fine-tuning of open-source models and as a prompting strategy in closed-source models, highlighting its adaptability and impact in GUI automation.

\textbf{Limitations.} For the benchmark, we have currently only collected primitive actions. To better approximate real-world scenarios, it is also important for the model to understand untrimmed videos. For the proposed framework, since the two stages are trained independently, the outcome of the GUI Action Narration depends on cursor grounding and keyframe extraction accuracy. Incorrect detections or hard-to-detect customized cursors can misguide the visual prompts, leading to erroneous action narrations.



\bibliographystyle{unsrt}

\bibliography{main}

\newpage

\appendix

\section{Appendix}

\subsection{Details of GUI cursor detection dataset}
\begin{figure}[htbp]
	\centering
\includegraphics[width=1\linewidth]{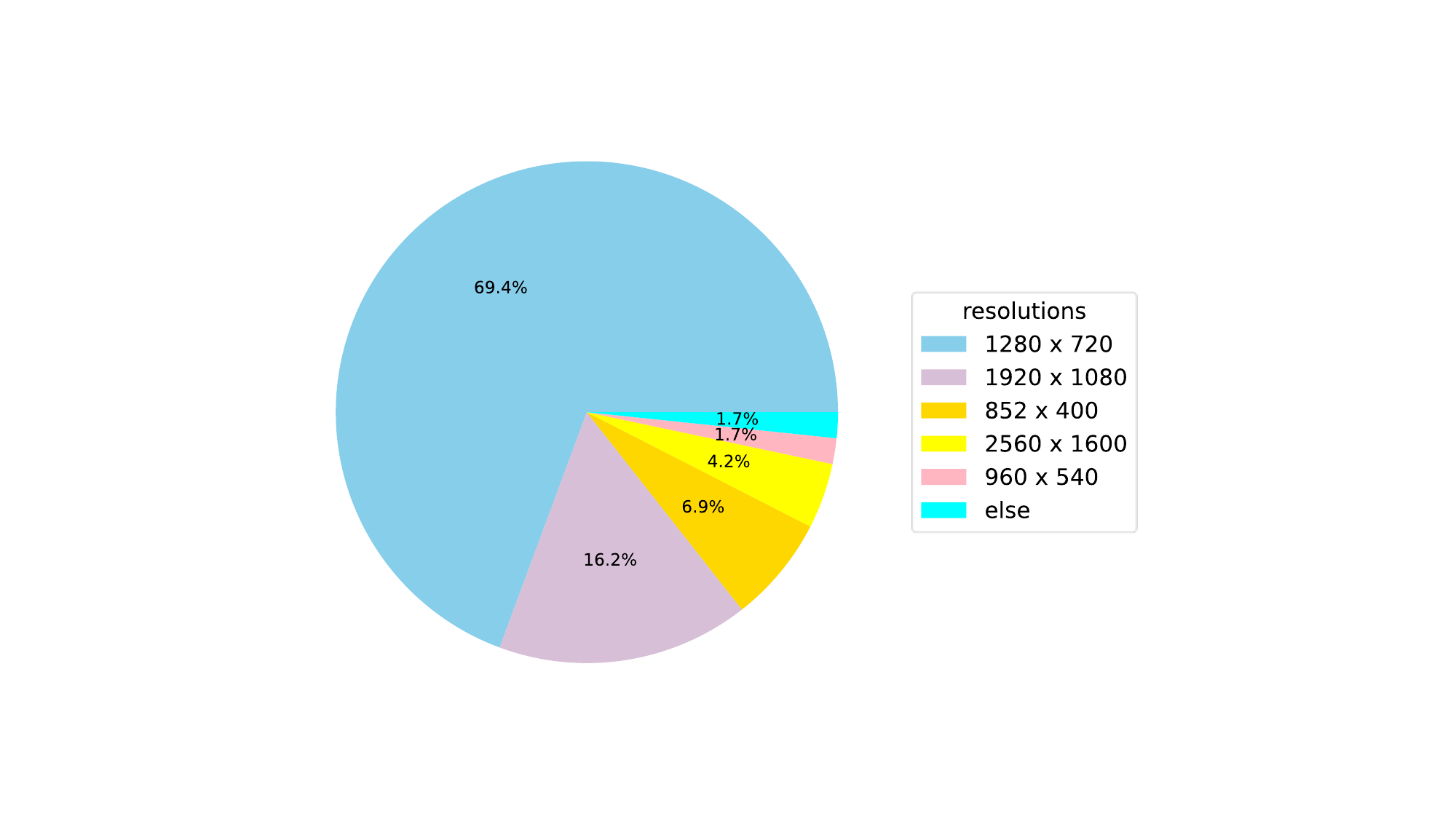}
\caption{\textbf{Distribution of the resolution of cursor datasets}: The main solution are 1280 $\times$ 720. Followed by the second largest population of 1920  $\times$ 720.}
\label{fig:cursor}
\end{figure}

The GUI cursor grounding dataset is in the format of screenshot coordinates pairs collected from tutorial videos under different resolutions with a scale of 6,027. Fig.~\ref{fig:cursor} shows the distribution of the screenshot resolution of the dataset.

\subsection{Further detail on GUI action narrator dataset}
All the videos in our dataset are sampled into 10 frames for training and testing. The following Fig.\ref{fig:frames} demonstrates the statistical distribution of GT-Keyframes in the videos. Start Frame and End Frame denote the index of the first GT-Keyframe and second GT-Keyframe. Duration (frames) are denoted as the number of frames between the first GT-Keyframes and the second GT-Keyframes. 
\begin{figure}[htbp]
	\centering
\includegraphics[width=1\linewidth]{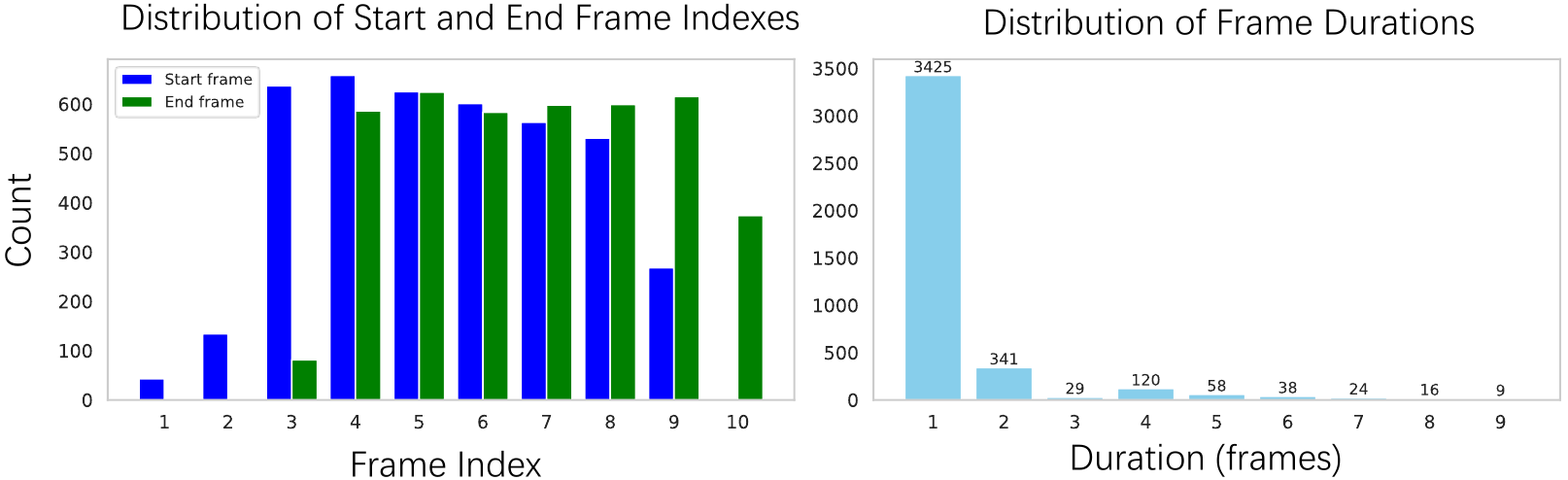}
\caption{Distribution of the Frames and Duration in our dataset. }
\label{fig:frames}

\end{figure}

\subsection{Details of Spatial detection model}
The detection model is trained on the YOLO-v8-ex for 100 epochs, and our best performance is achieved at the resolution 1536 $\times$ 834. Results are shown below in Fig.\ref{fig:cursor-res}.
\begin{figure}[htbp]
	\centering
\includegraphics[width=1\linewidth]{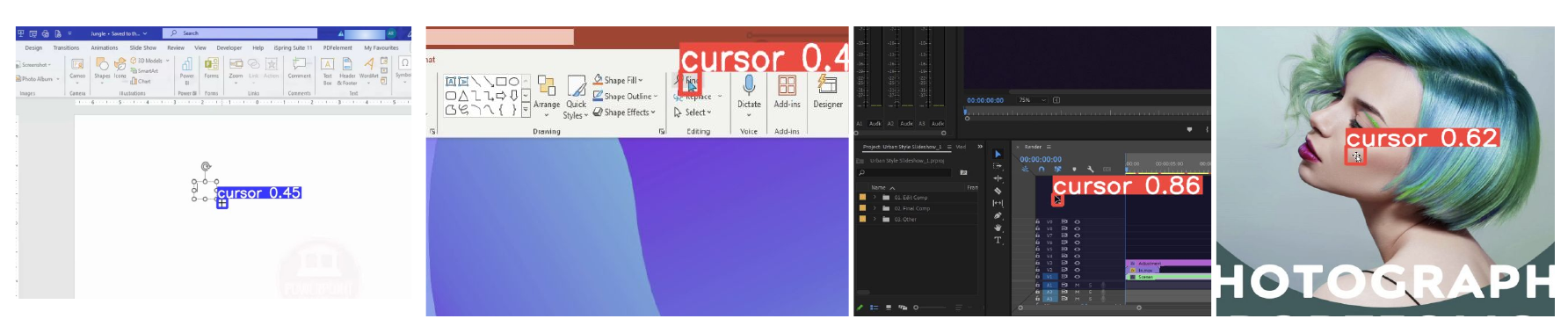}
\caption{\textbf{Cursor detection results}: The detection model can recognize cursors in different conditions}
\label{fig:cursor-res}
\end{figure}

\subsection{Quantitative result of the \modelname model}
\begin{figure}[htbp]
	\centering
\includegraphics[width=1\linewidth]{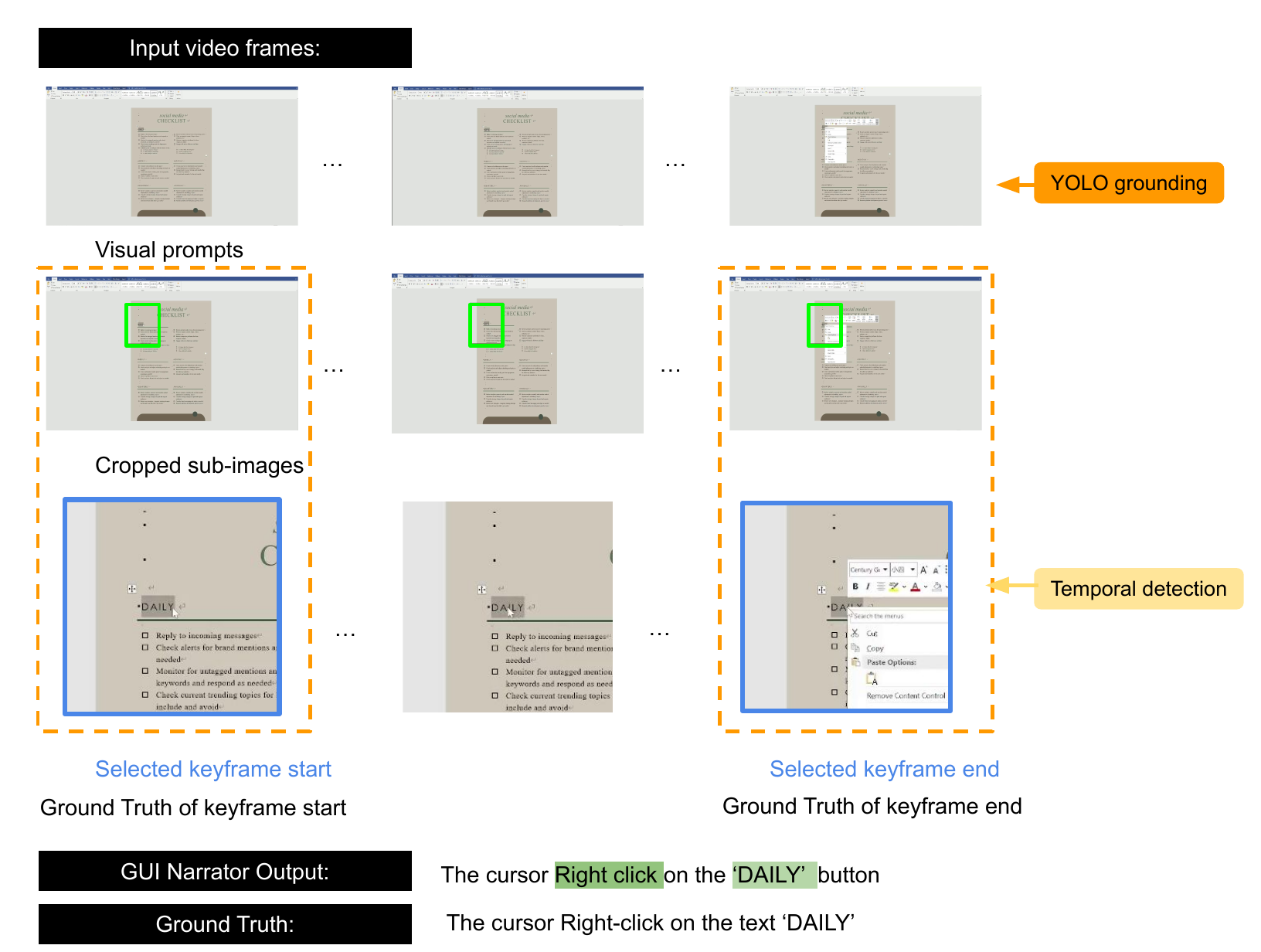}
\caption{\textbf{Success in atomic action narration of \modelname}: The YOLO grounding model and Temporal detection model accurately predicted the location of the cursor and keyframes, resulting in high-precision captioning for both Action type and Element.}
\label{fig:success_word}
\end{figure}

\begin{figure}[tp]
\centering
\vspace{-0.5cm}
\includegraphics[width=1\linewidth]{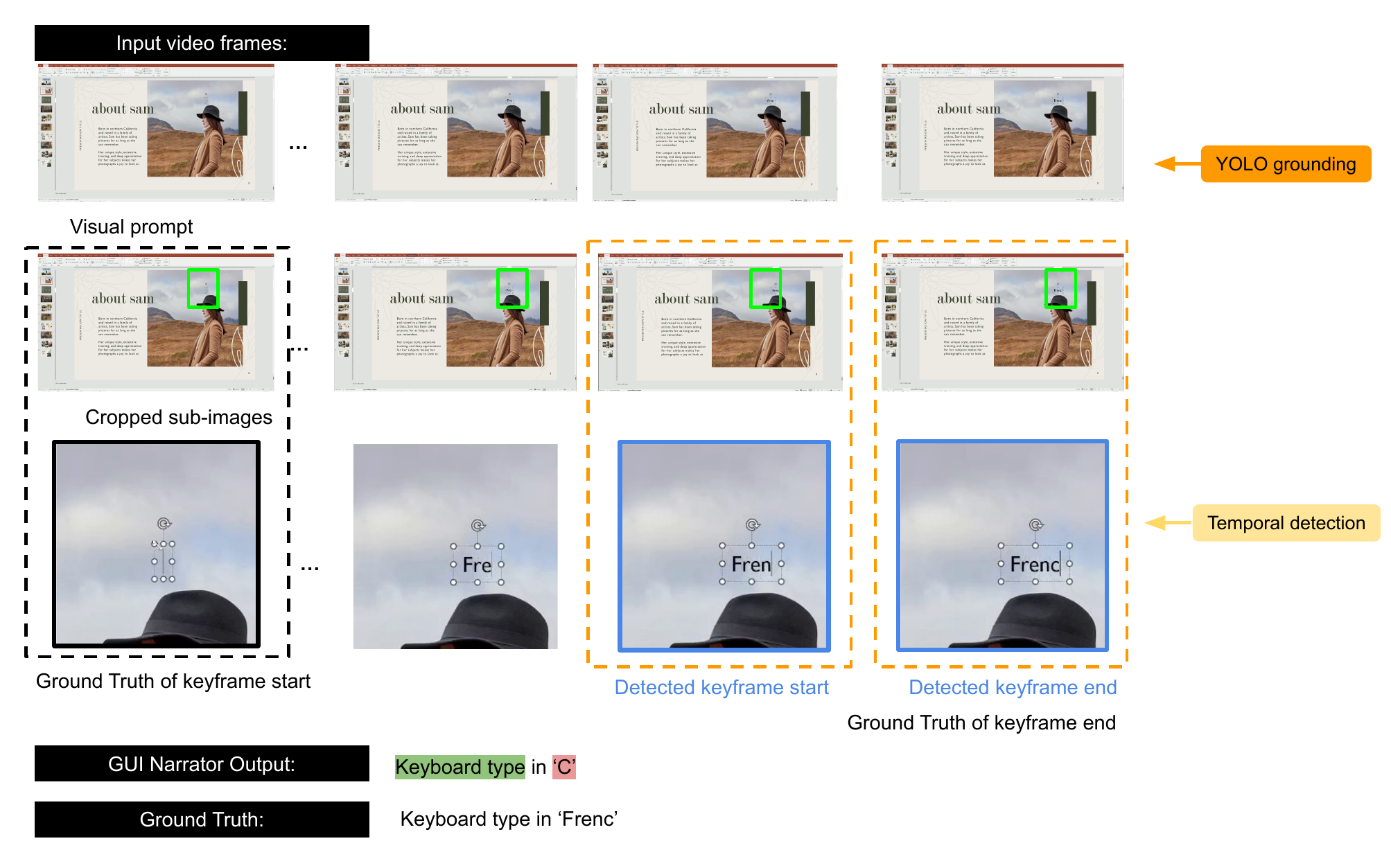}
\caption{\textbf{Failure in atomic action narration of \modelname}: The Temporal detection model predicted correctly on one of the key frames. It fails to predict the element. }
\label{fig:failure_ppt}
\end{figure}

We demonstrate our success and failure results on GUI layout of WORD and PowerPoint in the test dataset. Both GUI scenarios are not included in the GUI Narrator training dataset. Detailed results are shown in Fig.~\ref{fig:success_word} and Fig.~\ref{fig:failure_ppt} respectively.

\begin{figure}[bp]
\centering
\includegraphics[width=1\linewidth]{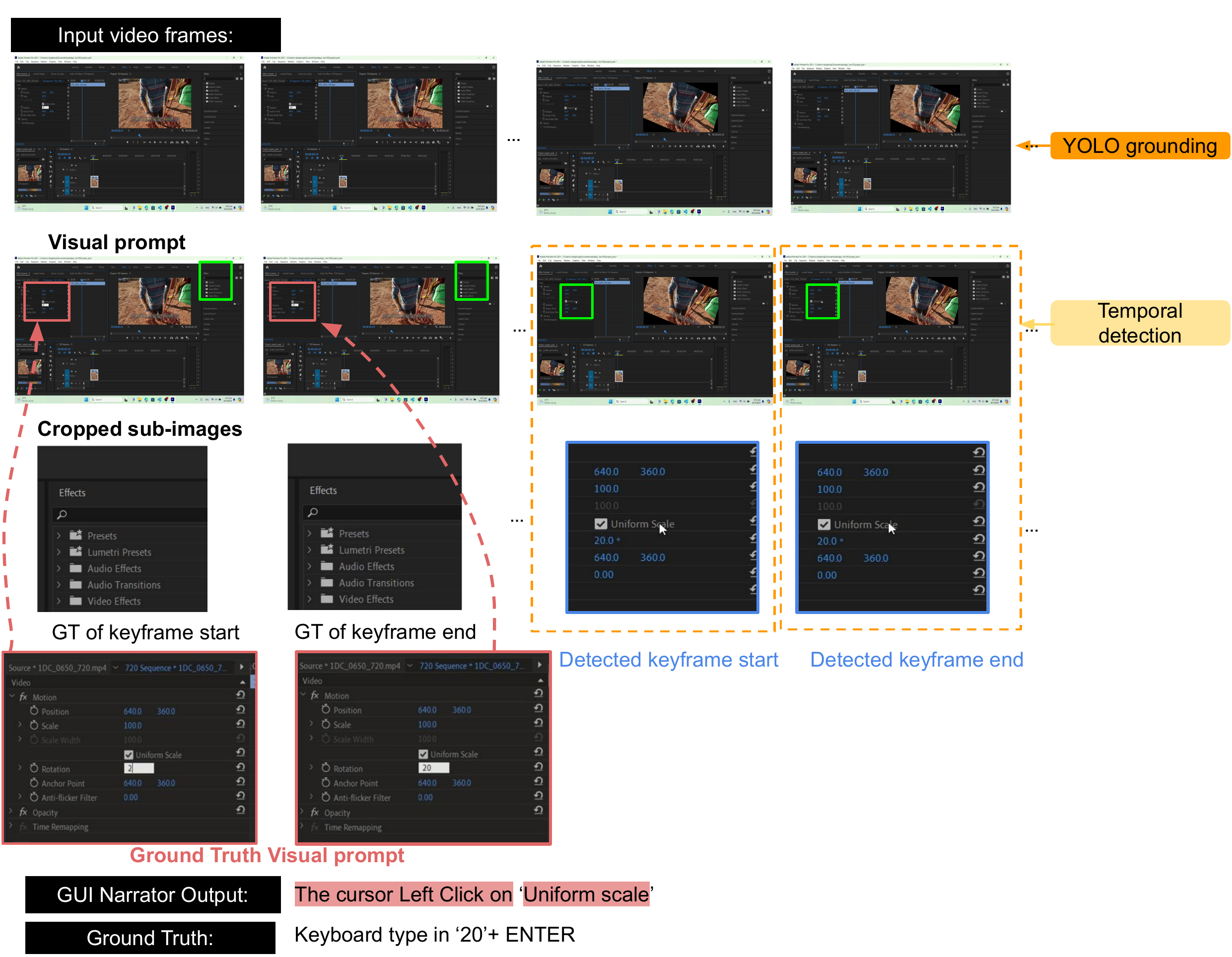}
\caption{\textbf{Failure in atomic action narration of \modelname}: The YOLO grounding model predicted incorrectly on the location of the cursor, leading to the failure in narration of VLM model. }
\label{fig:failure_ae}
\end{figure}

\newpage

\subsection{LLM evaluation prompt for evaluation in \datasetname benchmark}
\begin{figure}[htbp]
    \centering
    \begin{tcolorbox}[colback=gray!20, colframe=black, width=1\textwidth, arc=2mm, boxrule=0.5mm]
        \textbf{Evaluation prompt template} \\
        \\
        \textbf{\# Character Definition} \\
        You are an assistant to judge whether the given answer and the ground truth have the same Semantics meanings. ~\\ 
        ~\\
        \textbf{\# Guidelines} \\
    Action types are leftlick, rightclick, doubleclick, type write, drag.
    If the action is 'click' or 'keyboard type', split the description into [action type, element].
    If the action is 'Drag' split the description into [ action type, element, start(from), destination(to), purpose ]. ~\\
    Return the metric whether the each have the same semantic meaning: 0 for false, 1 for true.\\
    If the name of the element matches, the value will be 1.\\
    ~\\
    \textbf{\# Output Constraints } \\
    Only return a list 0 or 1 for each element in the format of [ , , , , ] for drag action and [ , ] for the click or type in actions. Don't provide the reason.\\
   ~\\  
     \textbf{\# Get started } \\
    The given ground truth: <\texttt{gt}>.
    The given answer: <\texttt{output}>.  ~\\
    \hrule 
    ~\\
    Assistant Justification:
    ~\\
    \end{tcolorbox}
    \caption{Evaluation prompt template}
\end{figure}

\end{document}